\pdfoutput=1

\documentclass[11pt]{article}
\usepackage{multirow} 
\usepackage{stfloats} 
\usepackage{enumitem}  
\usepackage{changepage}

\usepackage[preprint]{acl}

\usepackage{times}
\usepackage{latexsym}

\usepackage[T1]{fontenc}

\usepackage[utf8]{inputenc}

\usepackage{microtype}

\usepackage{inconsolata}

\usepackage{graphicx}

\title{The Elephant in the Coreference Room:\\Resolving Coreference in Full-Length French Fiction Works}

\author{Antoine Bourgois\and Thierry Poibeau\\
        Lattice (CNRS \& ENS-PSL \& Université Sorbonne Nouvelle), Montrouge, France\\
        \texttt{antoine.bourgois@protonmail.com}\\
        \texttt{thierry.poibeau@ens.psl.eu}
        }

\begin{document}
\maketitle
\begin{abstract}
While coreference resolution is attracting more interest than ever from computational literature researchers, representative datasets of fully annotated long documents remain surprisingly scarce. In this paper, we introduce a new annotated corpus of three full-length French novels, totaling over 285,000 tokens. Unlike previous datasets focused on shorter texts, our corpus addresses the challenges posed by long, complex literary works, enabling evaluation of coreference models in the context of long reference chains. We present a modular coreference resolution pipeline that allows for fine-grained error analysis. We show that our approach is competitive and scales effectively to long documents. Finally, we demonstrate its usefulness to infer the gender of fictional characters, showcasing its relevance for both literary analysis and downstream NLP tasks.
\end{abstract}

\section{Introduction}
Coreference Resolution (CR)—the task of identifying and grouping textual mentions that refer to the same entity (e.g., a person, an organization, a place)—is a fundamental component of natural language processing (NLP). It underpins downstream applications such as information extraction \citep{yao_2019}, text summarization \citep{liu_2021}, and machine translation \citep{vu_2024}. Over the past decades, significant progress has been made in CR, evolving from rule-based multi-sieve systems to end-to-end neural models, encoder-decoder architectures, and large language models based approaches, all contributing to improvements on benchmark datasets \citep{porada_2024}.

These models have long been trained and evaluated solely on generic datasets such as OntoNotes \citep{Hovy_2006}. As CR drew attention in other fields, it became evident that models trained on general datasets underperformed when applied to domain-specific tasks. To address this flaw, dedicated datasets have been developed, covering areas such as biomedical \citep{Lu_2021} and encyclopedic data \citep{Ghaddar_2016}.

Driven by the availability of extensive digitized collections, literary texts have emerged as a key subject of digital humanities \citep{Moretti_2013}. A large part of such research focuses on characters, considered a fundamental aspect of fiction works. The study of characters is essential for analyzing narrative structures, plot development or conducting diachronic studies. CR is crucial for applications such as quote attribution \citep{Vishnubhotla_2023}, character archetypes inference \citep{Bamman_2014}, and social networks extraction \citep{Elson_2010}. Additionally, it has been employed to study the representation and behavior of characters according to their gender \citep{Zundert_2023}.

As outlined by \citet{Roesiger_2018}, literary texts present unique challenges for CR, including character evolution throughout the narrative and the prevalence of dialogues involving multiple participants. They also contain a high proportion of pronouns and nested mentions. Complex narrative structures—such as letters, flashbacks, and sudden narrator interventions—further complicate the task. Additionally, authors often rely on readers' contextual understanding rather than explicit statements, creating ambiguities when linking mentions.

To address these challenges, annotated datasets have been developed, covering multiple languages and genres, from classical novels and fantasy tales to contemporary literature. These resources enable training and evaluating in-domain coreference resolution models, leading to steady performance improvements \citep{Martinelli_2024}. Despite visible progress on benchmarks, current state-of-the-art CR models still struggle with full-scale literary texts, limiting usefulness for downstream applications \citep{Vishnubhotla_2023}.

A key factor contributing to this limitation lies in the scarcity of fully annotated long documents. Most existing datasets consist of short excerpts or relatively brief texts. Since coreference annotation is labor-intensive and costly, there exists a trade-off between annotating a larger number of short documents or a smaller number of long ones. 

We argue that the lack of representative datasets for long literary texts is a major obstacle to effectively scaling CR models. This work aims to bridge this gap, and our contributions are as follows:
\begin{itemize}[leftmargin=10pt, itemsep=1pt, topsep=1pt]  
  \item an annotated dataset of character coreference for three full-length French novels spanning three centuries, showcasing the feasibility of combining automatic mention detection with manual coreference annotation.
  \item A modular CR pipeline scalable to long documents, enabling fine-grained error analysis and achieving competitive performance on benchmark dataset.
  \item A comprehensive study of the impact of document length on CR performance.
  \item A case study on character gender inference using CR models.\footnote{All code and data are publicly available at 
\href{https://github.com/lattice-8094/propp}{github.com/lattice-8094/propp}. 
The trained coreference resolution pipeline is readily usable through the open-source 
\href{https://pypi.org/project/propp-fr/}{\texttt{propp\_fr}} Python library.}.

\end{itemize}

\section{Related Work}

\subsection{Coreference Models}
Coreference resolution has undergone several paradigm shifts \citep{Poesio_2023}, evolving from rule-based, linguistically informed models tested on limited examples to data-driven statistical approaches enabled by the creation of large annotated datasets such as those from the Message Understanding Conference (MUC) and the Automatic Content Extraction (ACE) shared tasks \citep{Grishman_1995, Doddington_2004}.

The adoption of neural network-based models, beginning with \citet{Wiseman_2015}, marked significant progress. The introduction of end-to-end models by \citet{Lee_2017, Lee_2018}, further advanced CR by jointly detecting mention spans and resolving coreference, eliminating the need for external parsers and handcrafted mention detection models. Building on this foundation, higher-order inference (HOI) strategies and entity-level models were developed to refine entity representations during inference and leverage cluster-level information. 

However, as highlighted by \citet{Xu_2020}, the performance gains from these strategies have been marginal compared to the substantial improvements achieved by the use of more powerful encoders like ELMo, BERT and DeBERTaV3.

Alternative approaches using encoder-decoder architectures and large language models have been proposed, framing CR as sequence-to-sequence \citep{Hicke_2024} or question-answering \citep{Wu_2020, gan_2024} tasks. While showing promising results, these methods are computationally intensive and do not scale efficiently to long documents or resource-constrained scenarios.

\subsection{Existing Datasets}

\begin{table*}
\vspace{-1.5em}  
  \centering
\setlength{\tabcolsep}{3pt} 
  \begin{tabular}{@{}lcccccc@{}}
      \hline
      \multirow{2}{*}{\textbf{}} & \multirow{2}{*}{\textbf{Lang.}} & \multirow{2}{*}{\textbf{Domain}} & \multirow{2}{*}{\textbf{Doc.}} & \multirow{2}{*}{\textbf{Tokens}} & \multicolumn{2}{c}{\textbf{Tokens / Doc.}} \\
      \cline{6-7}
      & & & & & \textbf{Avg.} & \textbf{Max.} \\
      \hline
    \multicolumn{4}{@{}l@{}}{\textbf{Annotated Datasets}}\\
    OntoNotes\textsuperscript{en} \citep{Weischedel_2013} & English & Non-literary & 3,493 & 1,600,000 & 467 & 4,009 \\
    DROC \citep{Krug_2018} & German & Fiction & 90 & 393,164 & 4,368 & 15,718 \\
    RiddleCoref \citep{Cranenburgh_2019} & Dutch & Fiction & 21 & 107,143 & 5,102 & - \\
    LitBank \citep{Bamman_2020} & English & Fiction & 100 & 210,532 & 2,105 & 3,419 \\
    FantasyCoref \citep{Han_2021} & English & Fantasy & 214 & 367,891 & 1,719 & 13,471 \\
    KoCoNovel \citep{Kim_2024} & Korean & Fiction & 50 & 178,000 & 3,578 & 19,875 \\
    LitBank-fr \citep{Melanie_2024} & French & Fiction & 28 & 275,360 & 9,834 & 30,987 \\
    \hline
    \multicolumn{4}{@{}l@{}}{\textbf{Target Datasets}} \\
    Standard Ebooks\footnotemark & English & Fiction & 770 & 82,855,210 & 107,604 & 1,105,964 \\
    Chapitres \citep{Leblond_2022} & French & Fiction & 2,960 & 240,971,614 & 81,409 & 878,645 \\
    \hline
    \multicolumn{4}{@{}l@{}}{\textbf{Contribution}}\\
    Ours & French & Fiction & 3 & 285,176 & 95,058 & 115,415 \\
    \hline
  \end{tabular}
  \vspace{-0.5em}  
  \caption{\label{tab:datasets} Comparison of coreference annotation datasets: OntoNotes (English section), fiction datasets, and target datasets across languages.}
  \vspace{-1em}  
\end{table*}

While MUC and ACE laid the foundation for coreference datasets, OntoNotes has since become the primary benchmark for CR. Published in 2006 (\citeauthor{Hovy_2006}) and regularly updated, OntoNotes has been used in the CoNLL shared tasks \citep{Pradhan_2011, Pradhan_2012}. Its latest version \citep{Weischedel_2013} spans multiple languages (English, Chinese and Arabic), and genres, including conversations, news, web, and religious texts. The English part contains 1.6M tokens across 3,943 documents, averaging 467 tokens per document. OntoNotes does not contains singleton mentions—those that do not corefer with any other mention.

The growing interest for large literature corpora has driven the development of dedicated annotated datasets. The late 2010s saw the emergence of the first literary CR datasets, beginning with DROC \citep{Krug_2018}, including samples from 90 German novels annotated with character coreference chains. With over 393,000 tokens (averaging 4,368 tokens per document), DROC remains the largest literary CR dataset to date. The RiddleCoref dataset \citep{Cranenburgh_2019} followed, covering excerpts from 21 contemporary Dutch novels, though it is not publicly available due to copyright restrictions. \citet{Bamman_2020} released LitBank, consisting of the first 2,000 tokens from 100 English novels. This dataset covers six entity categories (persons, faculties, locations, geopolitical, organizations and vehicles). Other datasets include FantasyCoref \citep{Han_2021}, KoConovel covering 50 full-length Korean short stories \citep{Kim_2024}, and LitBank-fr \citep{Melanie_2024}. This last dataset is noteworthy in that it covers longer excerpts of text\textemdash{}averaging 9,834 tokens and up to 30,987 for the longest document.

\footnotetext{\href{https://standardebooks.org/subjects/fiction}{standardebooks.org}}

Despite these resources, extrinsic evaluations reveal that CR models perform poorly on full-length documents \citep{Zundert_2023}. Studies consistently show that performance degrades with increasing document length \citep{Joshi_2019, Toshniwal_2020, Shridhar_2023}. This represents a major challenge given that practical applications involve digitized collections such as Project Gutenberg or Wikisource, where documents frequently exceed 90,000 tokens and can reach up to a million as illustrated in Table~\ref{tab:datasets}.

While some initiatives annotate entire books, they often diverge from standard guidelines. \citet{He_2013} annotated \textit{Pride and Prejudice} but focused solely on proper mentions. Similarly, \citet{Zundert_2023} labeled character aliases across 170 novels, omitting pronouns and noun phrases. Other datasets, such as QuoteLi3 \citep{Muzny_2017} and PNDC \citep{Vishnubhotla_2022}, include coreference annotations for speakers and direct speech but lack broader character coverage. 

Until recently, the only coreference resolution results reported on a document of substantial length (37k tokens) came from \citet{Guo_2023}, though their work omits singletons, plural mentions, and nested entities. Since then, \citet{martinelli_2025} released an extended dataset, \textsc{BookCoref}\textsubscript{gold}, comprising two fully annotated English-language novels averaging 97{,}140 tokens per document, along with benchmark results, further illustrating the growing interest in long-document CR.

These observations underscore the need for an annotated corpus of full-length literary documents. Such a resource will enable more robust evaluation and improvement of CR models, addressing the gap between current datasets and intended applications.

\section{New Dataset}
We selected three average-length French novels spanning three centuries, resulting in a total of 285,176 tokens. We chose to annotate coreference for character mentions only for several reasons. First, most downstream tasks in literary NLP focus on characters. Second, previous work shows that characters account for the majority of annotated mentions\textemdash{}83.1\% in LitBank. Restricting annotations to character mentions allows us to leverage the 31,570 mentions already annotated in LitBank-fr to train an accurate mention detection model.

For consistency and interoperability, we adhere to the annotation guidelines from \citet{Melanie_2024}. We annotate all mentions referring to a character, including pronouns, nominal phrases, proper nouns, singletons and nested entities. Coreference links capture strict identity relations.

\vspace{1em}
\begin{adjustwidth}{0.5em}{0.5em}  
On [their]\textsubscript{1} way to visit [John]\textsubscript{2}, [[my]\textsubscript{3} parents]\textsubscript{1} met [[Mrs. Smith]\textsubscript{4} and [[her]\textsubscript{4} husband]\textsubscript{5}]\textsubscript{6}.
\end{adjustwidth}
\vspace{1em}

\noindent This sentence illustrates some annotation principles:
\begin{itemize}[leftmargin=10pt, itemsep=1pt, topsep=1pt]
  \item Mention types: pronoun (\textit{my}), nominal phrase (\textit{her husband}), and proper noun (\textit{John});
  \item Nested entities, including third-level nesting (e.g., \textit{her} within \textit{Mrs. Smith and her husband});
  \item Plural mentions (\textit{their}, \textit{my parents}, \textit{Mrs. Smith and her husband}) are treated as distinct coreference chains separate from their individual components;
  \item Singletons, such as \textit{John}, are annotated even if they are not referenced again.
\end{itemize}

\subsection{Mentions Detection Model}
\label{sec:mention_detection_model}
While \citet{Melanie_2024} report strong results for mention detection, we opted to retrain our own model. Our approach builds on a stacked BiLSTM-CRF architecture inspired by \citet{Ju_2018}, leveraging contextual token embeddings from CamemBERT\textsubscript{LARGE} \citep{Martin_2020}. When evaluating for exact match with gold annotations, We achieved an improvement of 4.99 in F1-score on the test set from LitBank-fr (Table~\ref{mention_performance}). To assess generalization performance and due to the small number of documents in the dataset, we also conducted a leave-one-out cross-validation (LOOCV). Details of the model architecture and hyperparameters are available in the Appendix \ref{sec:mention_detection_model_details}.

\begin{table}[htbp]
  \centering
  \setlength{\tabcolsep}{4pt} 
\renewcommand{\arraystretch}{1.1} 
  \begin{tabular}{@{}m{2.4cm}cccc@{}} 
    \hline
    \textbf{Model} & \textbf{P} & \textbf{R} & \textbf{F1} & \textbf{Support} \\
    \hline
    \citeauthor{Melanie_2024}\newline(test set) & 85.0 & 92.1 & 88.4 & 4,061 \\
    Ours (test set) & \textbf{91.29} & \textbf{95.59} & \textbf{93.39} & 4,061 \\
    \hline
    Ours (LOOCV) & 90.72 & 93.52 & 92.05 & 31,570 \\
    \hline
  \end{tabular}
  \vspace{-0.5em}  
  \caption{\label{mention_performance} Mention detection performances.}
  \vspace{-1em}  
\end{table}

Coreference annotation is usually carried out in two stages: annotating the mention spans, then linking mentions referring to the same entity together. Given our model's 92.05 F1-score, we consider its performance sufficient to automate the first operation, significantly reducing annotation time.

\subsection{Coreference Annotation}

Coreference annotation is performed manually, building on the automatically detected mentions. A single annotator reviews the text, assigns entity identifiers to each mention, corrects errors from the mention detection step, deleting spurious mentions, adding missed ones, and adjusting incorrect boundaries. This process yield gold-standard annotations for both mentions and coreference chains.

To assess annotation consistency, we double-annotated a sample from each of the three novels (5,000 tokens per text, ~5\% of the corpus). Inter-annotator agreement (IAA) was measured for mention spans (F1-score) and coreference chains (MUC, B\textsuperscript{3}, and CEAF\textsubscript{e}). Results show high consistency: mention span F1-score of 97.47 (vs. 86.0 in \citet{Bamman_2019}), benefiting from our focus on a single, well-defined entity type. Coreference agreement is also high: MUC 96.40, B\textsuperscript{3} 91.02, and CEAF\textsubscript{e} 71.65 (86.36 CoNLL F1). The lower CEAF\textsubscript{e} reflects differences in annotator decisions regarding long coreference chains and ambiguous cases such as plural entities leaving room for multiple valid interpretations. These results overall demonstrate the reliability and robustness of our annotations.

To perform annotation we use SACR, an open-source, browser-based interface \citep{Oberle_2018}. This tool meets our requirements, allowing efficient processing of long texts, tracking a large number of entities and handling nested mentions.

Mention detection errors mainly involve difficult cases, such as nested and ambiguous mentions (animals with agentivity, appositions, reflexive pronouns) or other edge cases. It shows the feasibility of leveraging automatic mention detection to accelerate coreference annotation. The manual annotation of a 100k-token text takes around 40 hours.

\subsection{Dataset Statistics}
Table~\ref{dataset_stats} summarizes statistics from our dataset. The entity spread refers to the distance between the first and the last mention of an entity \citep{Toshniwal_2020}. This highlights a key specificity of literary texts, characters can be referred to thousands times over several hundred pages, comprising thousands of tokens.

\begin{table}[htbp]
  \vspace{-0.5em}  
  \centering
  \setlength{\tabcolsep}{10pt} 
  \renewcommand{\arraystretch}{1} 
  \begin{tabular}{@{}lr@{}} 
    \hline
    Average Mentions / Doc. & 13,178 \\
    Singletons Ratio & 1.15\% \\
    Coreference Chains / Doc. & 159 \\
    Average Mentions / Chain & 82 \\
    Maximum Mentions / Chain & 4,932 \\
    Average Entity Spread (tokens) & 17,529 \\
    Maximum Entity Spread (tokens) & 115,369 \\
    Second-Level Nested Mentions & 5.74\% \\
    Third-Level Nested Mentions & 0.30\% \\
    Plural Mentions Ratio & 8.13\% \\
    Proper Mentions & 12.79\% \\
    Nominal Mentions & 12.26\% \\
    Pronominal Mentions & 74.95\% \\
    \hline
  \end{tabular}
  \vspace{-0.5em}  
  \caption{\label{dataset_stats} Dataset statistics summary.}
  \vspace{-1em}  
\end{table}

Another important metric for characterizing coreference is the distance to the nearest antecedent \citep{Han_2021}. For each mention, we locate the previous mention belonging to the same coreference chain and measure the difference in terms of mention positions. \citet{Bamman_2020} analyzed the distribution of distance to nearest antecedent for proper nouns, noun phrases and pronouns. We replicate their experiment and report similar results. While 95\% of pronouns appear within 7 mentions of their last antecedent, this distance reach up to 270 mentions for proper nouns and noun phrases. This observation calls for distinct handling of pronouns, common, and proper nouns during CR. The the last 1\% of proper and common noun mentions exhibit a distance of over 1,700 mentions, presenting a significant challenge for CR. See Appendix \ref{sec:antecedent_distribution} for the full distribution of antecedent distances. 

\subsection{Corpus Merging}
Since we followed the guidelines from \citet{Melanie_2024}, the newly annotated dataset is fully compatible with the character annotations from the LitBank-fr dataset. It allows us to merge the two datasets, resulting in a combined dataset containing 31 documents and 71,105 character mentions. This decision is motivated by the goal of evaluating generalization across a broader range of texts.

This merged dataset becomes the largest annotated literary coreference dataset in terms of tokens (560,536), average document length (18,081 tokens), and maximum document length (115,415 tokens). Unless otherwise specified, all results presented in this paper pertain to this merged corpus, which we refer to as \texttt{Long-LitBank-fr}.

\section{Coreference Resolution}
Several coreference resolution pipelines are available off-the-shelf, such as the \textit{CoreferenceResolver} module from Spacy\footnote{\href{https://spacy.io/api/coref}{https://spacy.io/api/coref}}, Fastcoref \citep{Otmazgin_2022} and AllenNLP \citep{Gardner_2018}. BookNLP \citep{Bamman_2020}, is a pipeline performing, among other, mentions detection and coreference resolution for English. A French adaptation, BookNLP-fr, was developed by \citet{Melanie_2024} and trained on the LitBank-fr dataset. The BookNLP pipelines implement an end-to-end coreference resolution model \citep{Ju_2018}.

Diverging from recent trends of end-to-end architectures, we propose to implement coreference resolution as a modular pipeline, facilitating the study of each component's role and enabling fine-grained error analysis.

Additionally, the use of compact, specialised models ($\sim$15M and $\sim$11M parameters for mention detection and mention scoring models) is motivated by practical end-use considerations: the need to process large literary corpora under limited computational resources. This is further supported by recent critiques of the "bigger-is-better" trend in AI, arguing that simply increasing scale doesn’t always lead to better results. Instead, smaller, task-specific models have been shown to offer more sustainable, transparent, and often competitive solutions for domain-specific applications \citep{Varoquaux_2025}.

\subsection{Pipeline Description}
Our mention-pair-based coreference resolution pipeline is composed of the following modules:
\\[0.5em]
\textbf{Mention Detection}: We employ the mention detection module described in section \ref{sec:mention_detection_model}, which consists of a stacked BiLSTM-CRF architecture using token-level embeddings from pretrained CamemBERT\textsubscript{LARGE} model as input. We retrained it on the merged corpus, achieving an increase of 2.82 points in F1-score (94.87). As mention detection can impact overall CR performance, we make it possible to bypass the errors introduced by this module by using gold mentions as input to the mention-pair encoder.
\\[0.5em]
\textbf{Considered Antecedents}: To address the quadratic complexity of considering all antecedents, recent approaches introduce hyperparameters to uniformly limit the number of considered antecedents \citep{Thirukovalluru_2021, Wu_2020}. Inspired by \citet{Bamman_2020} and supported by our observations regarding antecedent distance, we adopt a mention-type-specific approach. We limit the number of antecedents to 30 for pronouns and 300 for proper and common nouns. 
\\[0.5em]
\textbf{Mention Pair Encoder}: Mention-pairs are encoded by concatenating the representations of the two mentions with a feature vector that includes attributes such as gender, grammatical person, and the distance between the mentions. For multi-token mentions, the representation is calculated as the average of the first and last tokens embeddings.
\\[0.5em]
\textbf{Mention Pair Scorer}: Encoded mention-pairs are passed into a feedforward neural network trained to predict if two mentions refer to the same entity. Details about the features, model architecture and parameters are provided in the Appendix \ref{sec:mention_pairs_model}.
\\[0.5em]
\textbf{Antecedent Ranker}: Following \citet{Wiseman_2015}, candidate antecedents are ranked according to their predicted scores. During inference, the highest-scoring antecedent is selected unless all scores fall below 0.5, in which case the null antecedent is assigned.
\\[0.5em]
\textbf{Entity Clustering}: Default strategy for linking mentions into clusters is to scan the document from left to right, each new mention is either merged into the cluster of its best-ranked antecedent or left as a standalone entity. Coreference chains are defined as the set of mentions in a cluster. 

We explore additional strategies to address specific challenges and improve overall performance.
\\[0.5em]
\textbf{Handling Limited Antecedents}: Limiting the number of antecedents can lead to split coreference chains. A common strategy in literary texts is to link all matching proper nouns at the document level, along with their derivatives. While previous works have been using hand-crafted sets of aliases to link proper mentions \citep{Bamman_2020}, we leverage local mention-pairs scoring to perform coreference resolution at the document scale. Let's say that all local predictions involving mentions of "Sir Ralph Brown" and "Raphael" are coreferent, we propagate this decision to all mention-pairs at the global scale, bridging the gap between a mention and an antecedent that would otherwise be out of the range of locally considered antecedents.
\\[0.5em]
\textbf{Leveraging Non-Coreference Predictions}: While most mention-pair models focus on coreference links, the cross-entropy loss used during training involves that they are equally trained to predict non-coreference. We propose leveraging high-confidence non-coreference predictions to prevent later incorrect cluster merging. Mention-pairs containing a coordinating conjunction, such as “[Ralph] and [Mr. Delmare]”, are a strong indication of non-coreference between these mentions, which can be used to prevent the merging of these entities at document level. This approach is combined with an "easy-first" clustering strategy \citep{Clark_2016}, which processes mentions in order of confidence rather than left-to-right, thus delaying harder decisions.

The addition of these two strategies is refered to as the \textit{easy-first, global proper mentions coreference approach}. This approach follows a hierarchical iterative process, where high-confidence local mention-pair predictions are resolved first, constraining subsequent decisions at the document level. This post-processing module is not trained.

\subsection{Evaluation Metrics}

We evaluate CR performance using MUC \citep{Vilain_1995}, B\textsuperscript{3} \citep{Bagga_1998}, and CEAF\textsubscript{e} \citep{Luo_2005} scores. For overall performance assessment we report the average F1-score of the three metrics which we refer to as the CoNLL F1-score \citep{Pradhan_2012}. We use the scorer implementation by Grobol.\footnote{\href{https://github.com/LoicGrobol/scorch}{https://github.com/LoicGrobol/scorch}}

\subsection{Document Length}
While \citet{Poot_2020} investigated the impact of document length on CR by truncating documents to different sizes, we adopt a splitting approach. This allows us to evaluate CR performance on more text excerpts.

Given a target sample size of \( L \) tokens, we first select all documents from our corpus that exceed this length. Each document is split into non-overlapping samples, each containing \( L \) tokens. CR is performed independently on each sample, and the results are averaged across samples of a given document. The overall CR scores are calculated as the macro-average across all retained documents.

\subsection{Coreference Resolution Results}

\subsubsection{Mention-Pairs Scorer Results}
The mention-pairs scorer, evaluated using leave-one-out cross-validation with gold mention spans, achieved an overall accuracy of 88.10\%. As shown in Table~\ref{Mention_Pairs_Scorer}, performance disparities between classes reflect the underlying class imbalance, with significantly higher precision and recall for non-coreferent pairs (class 0). Most errors occurred for mention pairs where the scorer's confidence is low ($\sim$0.5) (Appendix \ref{sec:mention_pairs_errors}). As we use the highest ranked antecedent strategy, not all scorer decisions are used during entity clustering, mitigating the number of wrong decisions considered.

\begin{table}[htbp]
  \centering
  \setlength{\tabcolsep}{7pt} 
\renewcommand{\arraystretch}{1.2} 
  \begin{tabular}{@{}ccccc@{}} 
    \hline
    \textbf{Coref.} & \textbf{P} & \textbf{R} & \textbf{F1} & \textbf{Support} \\
    \hline
    \textbf{0} & 92.31 & 93.18 & 92.74 & 5.52M (82\%)\\
    \textbf{1} & 68.49 & 65.62 & 67.02 & 1.25M (18\%)\\
    \hline
  \end{tabular}
  \vspace{-0.5em}  
  \caption{\label{Mention_Pairs_Scorer}Mention-pairs scorer performance on Long-LitBank-fr corpus. Precision (P), Recall (R).}
  \vspace{-1.5em}  
\end{table}

\subsubsection{Highest Ranked Antecedent}
After sorting, the correct antecedent was predicted in 88.05\% of cases, highlighting the effectiveness of this approach. Errors occurred for 8,496 mentions (11.95\%). In 1,478 cases (2.08\%), the range of considered antecedents is too narrow, leaving true antecedents out of reach. For these mentions, the null antecedent is assigned approximately half the time, while an unrelated antecedent is assigned in the other half. In 7,018 cases (9.87\%), the true antecedent is within reach, but the model incorrectly assigned a different antecedent in nearly 90\% of instances. In the remaining 10\%, the null antecedent is wrongly predicted.

The additional global proper mentions coreference strategy aims at reducing both types of errors, by bridging the gap between proper mentions and their long distance antecedent, and by limiting clustering of mentions that are believed to be distinct from local mention-pair scores.

\begin{figure*}[tp]
\vspace{-1.5em}  
  \centering
  \includegraphics[width=\textwidth]{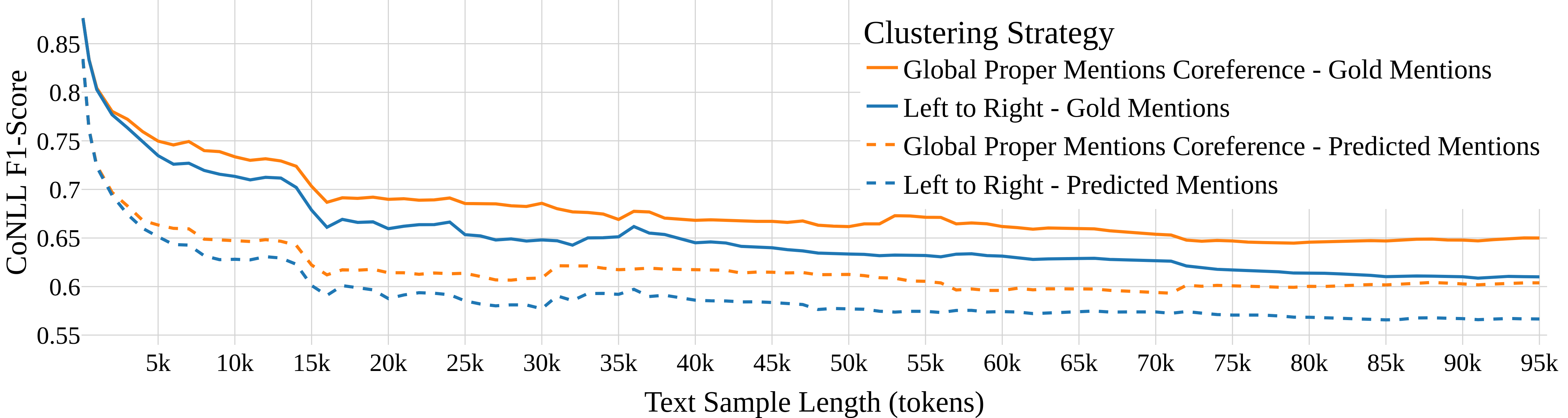}  
  \vspace{-1.5em}  
  \caption{Impact of document length on CR performance for different strategy. Gold and predicted mentions.}
  \vspace{-1.em}  
  \label{fig:document_length_impact}
\end{figure*}

\begin{table*}[bp]
  \vspace{-0.5em}  
  \centering
  \setlength{\tabcolsep}{0pt} 
  \renewcommand{\arraystretch}{1.1} 
  \small
  {\fontsize{9.5}{12}\selectfont 
  \begin{tabular}{@{}p{3.7cm} >{\centering\arraybackslash}p{3.0cm} >{\centering\arraybackslash}p{1.6cm} >{\centering\arraybackslash}p{2.1cm} >{\centering\arraybackslash}p{1.3cm} >{\centering\arraybackslash}p{1.3cm} >{\centering\arraybackslash}p{1.3cm}>{\centering\arraybackslash}p{1.3cm}@{}} 
  \hline
    \textbf{Corpus (test set)} & \textbf{Model} & \textbf{Mentions} & \textbf{Tokens / Doc} & \textbf{MUC} & \textbf{B\textsuperscript{3}} & \textbf{CEAFe} & \textbf{CoNLL} \\
    \hline
    LitBank-fr (test-set) & \citealt{Melanie_2024} & Gold & 2,000 & 88.0 & 69.2 & 71.8 & 76.4 \\
    LitBank-fr (test-set) & Ours & Gold & 2,000 & \textbf{92.43} & \textbf{70.67} & \textbf{75.59} & \textbf{79.56} \\
    \hline
    \hline
    Long-LitBank-fr (3 docs) & Ours & Gold & 93,019 & 96.64 & 52.36 & 46.45 & 65.15 \\
    Long-LitBank-fr (3 docs) & Ours & Predicted & 93,019 & 95.59 & 45.4 & 35.95 & 58.98 \\
    \hline
  \end{tabular}}
  \vspace{-0.5em}
  \caption{\label{entity_comparison}CR performance on LitBank-fr test-set and on the three fully annotated novels. Gold and predicted mentions.}
  \vspace{-2.5em}  
\end{table*}

\subsubsection{Entity Clustering Strategies}
The global proper mentions strategy leads to an overall gain in performance measured by CoNLL F1-score of 1.68 points. We observe a slight drop for MUC, but a significant improvement on both B\textsuperscript{3} and CEAF\textsubscript{e}.
\begin{table}[htbp]
  \vspace{-0.5em}  
  \centering
  \setlength{\tabcolsep}{1pt} 
    {\fontsize{10.5}{12}\selectfont 
  \renewcommand{\arraystretch}{1.2} 
  \begin{tabular}{@{}m{2.7cm} >{\centering\arraybackslash}m{1.1cm} >{\centering\arraybackslash}m{1.1cm} >{\centering\arraybackslash}m{1.3cm} >{\centering\arraybackslash}m{1.2cm}@{}}
    \hline
    \textbf{Strategy} & \textbf{MUC} & \textbf{B\textsuperscript{3}} & \textbf{CEAF\textsubscript{e}} & \textbf{CoNLL} \\
    \hline
    Left to Right & \textbf{94.61} & 62.95 & 60.36 & 72.64\\
    Global Proper CR & 94.45 & \textbf{67.32} & \textbf{61.18} & \textbf{74.32}\\
    \hline
  \end{tabular}}
  \vspace{-0.5em}  
  \caption{\label{entity_clustering_strategies}Coreference resolution for Long-LitBank-fr corpus. Average F1-scores. Gold mentions.}
  \vspace{-0.75em}  
\end{table}

These scores reflect the overall performance gain of this strategy on the full \texttt{Long-LitBank-fr} corpus (averaging 18,081 tokens per document). However, it is best suited to long texts that present both the risk of out-of-reach antecedent, and sufficient local evidence on proper mentions-pairs to propagate document-wide decisions. 

\subsubsection{Influence of Document Length}
When analyzing performance gains as a function of document length, we observe that the MUC score remains relatively stable. For CEAF\textsubscript{e} we see a consistent improvement of around 1 point, regardless of document length. The most striking trend is observed on the B\textsuperscript{3} score: for documents exceeding 20,000 tokens, the gain from the global proper mentions strategy increases significantly, ranging from 5 to 10 points. See Appendix \ref{sec:absolute_gain_clustering_strategy}.

From Figure~\ref{fig:document_length_impact}, we observe that the overall CR performance decreases with document length. Much of the performance loss is observed in the lower range. This might well explain why CR models trained and evaluated on documents of limited length (<10k), have been deceiving when used for downstream tasks on full length documents.

The proper mentions global coreference strategy consistently outperform the vanilla left-to-right method. Performance gains is mostly negligible for short documents (< 2k tokens), but becomes significant and stable beyond, reaching +3 points on the CoNLL F1-score. This shows the effectiveness of our approach for handling CR in longer documents.

Additionally, Figure~\ref{fig:document_length_impact} shows the impact of using predicted mentions as input to the mention-pair encoder, leading to a performance drop of $\sim$7\%, this result is consistent with previous publications.

\subsubsection{Comparison to Baseline}
For French, our new pipeline consistently outperforms the model proposed by \citet{Melanie_2024} on their test set, setting a new baseline on this specific dataset. We also report average performances on the 3 newly annotated novels for future comparison ; both with gold and predicted mentions.

See Appendix \ref{sec:coref_benchmarks_comparison} for cross-dataset and cross-language coreference performance comparison.


While this experiment reveals performance limitations exacerbated by document length, commonly used CR metrics (MUC, B\textsuperscript{3}, CEAF\textsubscript{e}) have been criticised for presenting systematic flaws. Alternative metrics such as LEA \citep{Moosavi_2016} and BLANC \citep{Recasens_2011} have been proposed as better aligned with linguistic intuitions. Others argue for extrinsic evaluation \citep{OKeefe_2013, Vishnubhotla_2023}, where CR is assessed based on its contribution to easier to evaluate, downstream tasks.

\section{Gender Prediction Case study}
As mentioned, studies gravitating around character gender have attracted substantial attention from computational humanities researchers \citep{Underwood_2018}. A key challenge is accurately predicting the gender of as many character mentions as possible to ensure representative results.

Early works relied on heuristics to infer gender from explicit clues (he, Mrs, the man), achieving high precision (90\%) but lower recall (30-50\%), due to the high proportion of ambiguous mentions in literary texts. Recent works leverages CR for broader gender prediction \citep{Vianne_2023}.

\subsection{Data Preparation}
We use the \textit{Long-Litbank-fr} corpus. Starting with all character mentions, we discard singletons (2.74\%) and plural mentions (9.84\%). We manually annotate the gender of the remaining 62,162 mentions at the entity level. We adopt a binary approach to gender. Works of fiction are subject to play on characters' gender, such as gender revelation or asymmetry of knowledge between characters. To assign character gender we adopt the omniscient perspective \citep{Kim_2024}, refering to the knowledge one have at the end of the entire book. We discard chains whose gender cannot be annotated with certainty, leaving us with 804 entities and 61,852 mentions (86.99\% of all mentions).

\subsection{Prediction Pipeline}

To predict the gender of character mentions we implement a multi-stage solution:
\\[0.4em]
\textbf{Heuristic rules}: assign gender based on heuristics from explicit gender clues (pronouns, noun phrases, articles and adjectives).
\\[0.4em]
\textbf{First-name database}: determine the gender of proper mentions using a statistical database of first names given in France since 1900.\footnote{\href{https://www.insee.fr/fr/statistiques/8205621?sommaire=8205628}{French National Institute of Statistics and Economic Studies (\textit{INSEE}).}}
\\[0.4em]
\textbf{Coreference propagation}: resolve coreference, compute the male/female ratio of processed mentions, and assign the majority gender to all mentions within the coreference chain.

We compare our results with those of \citet{Naguib_2022} who used a similar combination of heuristic rules and CR to infer character gender.

\subsection{Case Study Results}
CR significantly improves recall compared to rule-based methods. While heuristics achieve high precision (>98\%), they suffer from low recall (37-47\%), reflecting the significant number of mentions whose gender cannot be inferred without additional context. Our approach outperforms the baseline by leveraging sophisticated heuristic rules, a first-names database, and a more effective CR pipeline. Although CR slightly reduces precision—a consequence of clustering errors—the substantial recall gain makes it a robust method overall.

\begin{table}[htbp]
  \centering
  \footnotesize
  \setlength{\tabcolsep}{2pt} 
  \renewcommand{\arraystretch}{1.2} 
  \begin{tabular}{@{}m{2.35cm}
    >{\centering\arraybackslash}m{0.7cm}
    >{\centering\arraybackslash}m{0.7cm}
    >{\centering\arraybackslash}m{0.7cm}
    >{\centering\arraybackslash}m{0.7cm}
    >{\centering\arraybackslash}m{0.7cm}
    >{\centering\arraybackslash}m{0.7cm}@{}}
    \hline
    & \multicolumn{3}{c}{\textbf{Male}} & \multicolumn{3}{c}{\textbf{Female}} \\
    \cline{2-7}
    & \textbf{P} & \textbf{R} & \textbf{F1} & \textbf{P} & \textbf{R} & \textbf{F1} \\
    \hline
    Baseline\newline\citeauthor{Naguib_2022} \citeyear{Naguib_2022} & 95.0 & 45.0 & 60.6 & 97.0 & 58.0 & 72.7 \\
    \hline
    Heuristic Rules & \textbf{99.8} & 37.0 & 54.0 & \textbf{98.9} & 46.7 & 63.4 \\
    + First-name data & 99.8 & 38.4 & 55.4 & 98.8 & 47.4 & 64.1 \\
    + Coreference & 95.4 & \textbf{91.6} & \textbf{93.4} & 90.4 & \textbf{93.4} & \textbf{91.9} \\
    \hline
  \end{tabular}
  \vspace{-0.5em} 
  \caption{\label{Gender_Study_Case}Mentions gender prediction performance (Precision, Recall, F1).}
  \vspace{-1.75em} 
\end{table}

\section{Conclusion}

We highlight critical limitations in coreference resolution (CR) for literary texts, particularly the scarcity of representative datasets, limiting the possibility to train and evaluate models tailored for literary computational studies. To bridge this gap, we release an annotated corpus of character coreference chains for three full-length French novels spanning three centuries (285,000+ tokens). We introduce a modular CR pipeline tailored for long documents, integrating global coreference propagation for proper nouns and an easy-first clustering approach. After carrying out a detailed error analysis of each component, we study the impact of document length on overall coreference performance. Our approach is competitive with existing state-of-the-art models, demonstrating good performance on longer texts. To demonstrate practical value, we apply it to character gender inference, significantly improving recall over rule-based baselines while maintaining high precision, and outperforming other CR-based approach. This study underscores the need for robust datasets and well-evaluated models to advance literary CR research.

\section*{Limitations}

While our dataset is among the largest annotated literary datasets in terms of tokens (285,000), it is limited by the fact that it only contains three documents. This implies that it does not encompass the full diversity of time periods, literary movements, and genres within French literature. This limitation may impact the generalizability of the coreference resolution (CR) models trained on this dataset. The proposed \textit{Long-LitBank-fr} corpus resulting from the concatenation with the \textit{LitBank-fr} dataset mitigates this issue by increasing diversity and improving the potential for model generalization. 

Another limitation is that we focused solely on annotating coreference chains for characters. Some downstream applications may require resolving coreference for other entity types (e.g., geographical entities, events). Since our annotations are restricted to characters, a model trained exclusively on this data may not easily transfer to tasks involving other entity types. In such cases, enriching the annotations would be necessary for broader applicability.

Furthermore, our study is limited to French-language texts, and we did not explore cross-lingual generalization of our pipeline. Expanding the dataset to include full documents in other languages could improve its applicability. This could be achieved through annotation transfer or by leveraging multilingual models, which would help reduce the cost of manual annotation. 

Finally, while extrinsic evaluation is not the primary focus of this work, we have only begun to assess our pipeline through its application to character gender inference. A more comprehensive evaluation of the models' suitability for full-document literary analysis would require additional extrinsic assessments, such as network extraction or quote attribution.






\newpage
\bibliography{custom}

\begin{thebibliography}{61}
\providecommand{\natexlab}[1]{#1}

\bibitem[{Bagga and Baldwin(1998)}]{Bagga_1998}
Amit Bagga and Breck Baldwin. 1998.
\newblock \href {https://api.semanticscholar.org/CorpusID:8622546}
  {Entity-based cross-document coreferencing using the vector space model}.
\newblock In \emph{International Conference on Computational Linguistics}.

\bibitem[{Bamman et~al.(2020)Bamman, Lewke, and Mansoor}]{Bamman_2020}
David Bamman, Olivia Lewke, and Anya Mansoor. 2020.
\newblock \href {https://aclanthology.org/2020.lrec-1.6/} {An annotated dataset
  of coreference in {E}nglish literature}.
\newblock In \emph{Proceedings of the Twelfth Language Resources and Evaluation
  Conference}, pages 44--54, Marseille, France. European Language Resources
  Association.

\bibitem[{Bamman et~al.(2019)Bamman, Popat, and Shen}]{Bamman_2019}
David Bamman, Sejal Popat, and Sheng Shen. 2019.
\newblock \href {https://doi.org/10.18653/v1/N19-1220} {An annotated dataset of
  literary entities}.
\newblock In \emph{Proceedings of the 2019 Conference of the North {A}merican
  Chapter of the Association for Computational Linguistics: Human Language
  Technologies, Volume 1 (Long and Short Papers)}, pages 2138--2144,
  Minneapolis, Minnesota. Association for Computational Linguistics.

\bibitem[{Bamman et~al.(2014)Bamman, Underwood, and Smith}]{Bamman_2014}
David Bamman, Ted Underwood, and Noah~A. Smith. 2014.
\newblock \href {https://doi.org/10.3115/v1/P14-1035} {A {B}ayesian mixed
  effects model of literary character}.
\newblock In \emph{Proceedings of the 52nd Annual Meeting of the Association
  for Computational Linguistics (Volume 1: Long Papers)}, pages 370--379,
  Baltimore, Maryland. Association for Computational Linguistics.

\bibitem[{Clark and Manning(2016)}]{Clark_2016}
Kevin Clark and Christopher~D. Manning. 2016.
\newblock \href {https://doi.org/10.18653/v1/P16-1061} {Improving coreference
  resolution by learning entity-level distributed representations}.
\newblock In \emph{Proceedings of the 54th Annual Meeting of the Association
  for Computational Linguistics (Volume 1: Long Papers)}, pages 643--653,
  Berlin, Germany. Association for Computational Linguistics.

\bibitem[{Doddington et~al.(2004)Doddington, Mitchell, Przybocki, Ramshaw,
  Strassel, and Weischedel}]{Doddington_2004}
George Doddington, Alexis Mitchell, Mark Przybocki, Lance Ramshaw, Stephanie
  Strassel, and Ralph Weischedel. 2004.
\newblock \href {https://aclanthology.org/L04-1011/} {The automatic content
  extraction ({ACE}) program {--} tasks, data, and evaluation}.
\newblock In \emph{Proceedings of the Fourth International Conference on
  Language Resources and Evaluation ({LREC}`04)}, Lisbon, Portugal. European
  Language Resources Association (ELRA).

\bibitem[{Elson et~al.(2010)Elson, Dames, and McKeown}]{Elson_2010}
David Elson, Nicholas Dames, and Kathleen McKeown. 2010.
\newblock \href {https://aclanthology.org/P10-1015/} {Extracting social
  networks from literary fiction}.
\newblock In \emph{Proceedings of the 48th Annual Meeting of the Association
  for Computational Linguistics}, pages 138--147, Uppsala, Sweden. Association
  for Computational Linguistics.

\bibitem[{Gan et~al.(2024)Gan, Poesio, and Yu}]{gan_2024}
Yujian Gan, Massimo Poesio, and Juntao Yu. 2024.
\newblock \href {https://aclanthology.org/2024.lrec-main.145/} {Assessing the
  capabilities of large language models in coreference: An evaluation}.
\newblock In \emph{Proceedings of the 2024 Joint International Conference on
  Computational Linguistics, Language Resources and Evaluation (LREC-COLING
  2024)}, pages 1645--1665, Torino, Italia. ELRA and ICCL.

\bibitem[{Gardner et~al.(2018)Gardner, Grus, Neumann, Tafjord, Dasigi, Liu,
  Peters, Schmitz, and Zettlemoyer}]{Gardner_2018}
Matt Gardner, Joel Grus, Mark Neumann, Oyvind Tafjord, Pradeep Dasigi,
  Nelson~F. Liu, Matthew Peters, Michael Schmitz, and Luke Zettlemoyer. 2018.
\newblock \href {https://doi.org/10.18653/v1/W18-2501} {{A}llen{NLP}: A deep
  semantic natural language processing platform}.
\newblock In \emph{Proceedings of Workshop for {NLP} Open Source Software
  ({NLP}-{OSS})}, pages 1--6, Melbourne, Australia. Association for
  Computational Linguistics.

\bibitem[{Ghaddar and Langlais(2016)}]{Ghaddar_2016}
Abbas Ghaddar and Phillippe Langlais. 2016.
\newblock \href {https://aclanthology.org/L16-1021/} {{W}iki{C}oref: An
  {E}nglish coreference-annotated corpus of {W}ikipedia articles}.
\newblock In \emph{Proceedings of the Tenth International Conference on
  Language Resources and Evaluation ({LREC}`16)}, pages 136--142,
  Portoro{\v{z}}, Slovenia. European Language Resources Association (ELRA).

\bibitem[{Grishman and Sundheim(1995)}]{Grishman_1995}
Ralph Grishman and Beth Sundheim. 1995.
\newblock \href {https://aclanthology.org/M95-1001/} {Design of the {MUC}-6
  evaluation}.
\newblock In \emph{Sixth Message Understanding Conference ({MUC}-6):
  Proceedings of a Conference Held in {C}olumbia, {M}aryland, November 6-8,
  1995}.

\bibitem[{Guo et~al.(2023)Guo, Hu, Zhang, Qiu, and Zhang}]{Guo_2023}
Qipeng Guo, Xiangkun Hu, Yue Zhang, Xipeng Qiu, and Zheng Zhang. 2023.
\newblock \href {https://doi.org/10.18653/v1/2023.acl-long.851} {Dual cache for
  long document neural coreference resolution}.
\newblock In \emph{Proceedings of the 61st Annual Meeting of the Association
  for Computational Linguistics (Volume 1: Long Papers)}, pages 15272--15285,
  Toronto, Canada. Association for Computational Linguistics.

\bibitem[{Han et~al.(2021)Han, Seo, Kang, Kim, Choi, Song, and Choi}]{Han_2021}
Sooyoun Han, Sumin Seo, Minji Kang, Jongin Kim, Nayoung Choi, Min Song, and
  Jinho~D. Choi. 2021.
\newblock \href {https://doi.org/10.18653/v1/2021.crac-1.3} {{F}antasy{C}oref:
  Coreference resolution on fantasy literature through omniscient writer`s
  point of view}.
\newblock In \emph{Proceedings of the Fourth Workshop on Computational Models
  of Reference, Anaphora and Coreference}, pages 24--35, Punta Cana, Dominican
  Republic. Association for Computational Linguistics.

\bibitem[{He et~al.(2013)He, Barbosa, and Kondrak}]{He_2013}
Hua He, Denilson Barbosa, and Grzegorz Kondrak. 2013.
\newblock \href {https://aclanthology.org/P13-1129/} {Identification of
  speakers in novels}.
\newblock In \emph{Proceedings of the 51st Annual Meeting of the Association
  for Computational Linguistics (Volume 1: Long Papers)}, pages 1312--1320,
  Sofia, Bulgaria. Association for Computational Linguistics.

\bibitem[{Hicke and Mimno(2024)}]{Hicke_2024}
Rebecca Hicke and David Mimno. 2024.
\newblock \href {https://aclanthology.org/2024.latechclfl-1.27/} {{[Lions: 1]}
  and {[Tigers: 2]} and {[Bears: 3]}, oh my! literary coreference annotation
  with {LLM}s}.
\newblock In \emph{Proceedings of the 8th Joint SIGHUM Workshop on
  Computational Linguistics for Cultural Heritage, Social Sciences, Humanities
  and Literature (LaTeCH-CLfL 2024)}, pages 270--277, St. Julians, Malta.
  Association for Computational Linguistics.

\bibitem[{Hovy et~al.(2006)Hovy, Marcus, Palmer, Ramshaw, and
  Weischedel}]{Hovy_2006}
Eduard Hovy, Mitchell Marcus, Martha Palmer, Lance Ramshaw, and Ralph
  Weischedel. 2006.
\newblock \href {https://aclanthology.org/N06-2015/} {{O}nto{N}otes: The 90{\%}
  solution}.
\newblock In \emph{Proceedings of the Human Language Technology Conference of
  the {NAACL}, Companion Volume: Short Papers}, pages 57--60, New York City,
  USA. Association for Computational Linguistics.

\bibitem[{Joshi et~al.(2019)Joshi, Levy, Zettlemoyer, and Weld}]{Joshi_2019}
Mandar Joshi, Omer Levy, Luke Zettlemoyer, and Daniel Weld. 2019.
\newblock \href {https://doi.org/10.18653/v1/D19-1588} {{BERT} for coreference
  resolution: Baselines and analysis}.
\newblock In \emph{Proceedings of the 2019 Conference on Empirical Methods in
  Natural Language Processing and the 9th International Joint Conference on
  Natural Language Processing (EMNLP-IJCNLP)}, pages 5803--5808, Hong Kong,
  China. Association for Computational Linguistics.

\bibitem[{Ju et~al.(2018)Ju, Miwa, and Ananiadou}]{Ju_2018}
Meizhi Ju, Makoto Miwa, and Sophia Ananiadou. 2018.
\newblock \href {https://doi.org/10.18653/v1/N18-1131} {A neural layered model
  for nested named entity recognition}.
\newblock In \emph{Proceedings of the 2018 Conference of the North {A}merican
  Chapter of the Association for Computational Linguistics: Human Language
  Technologies, Volume 1 (Long Papers)}, pages 1446--1459, New Orleans,
  Louisiana. Association for Computational Linguistics.

\bibitem[{Kim et~al.(2024)Kim, Lee, and Lee}]{Kim_2024}
Kyuhee Kim, Surin Lee, and Sangah Lee. 2024.
\newblock \href {https://arxiv.org/abs/arXiv:2404.01140} {Koconovel: Annotated
  dataset of character coreference in korean novels}.

\bibitem[{Krug et~al.(2018)Krug, Weimer, Reger, Macharowsky, Feldhaus, Puppe,
  and Jannidis}]{Krug_2018}
Markus Krug, Lukas Weimer, Isabella Reger, Luisa Macharowsky, Stephan Feldhaus,
  Frank Puppe, and Fotis Jannidis. 2018.
\newblock \href {http://webdoc.sub.gwdg.de/pub/mon/dariah-de/dwp-2018-27.pdf}
  {Description of a corpus of character references in german novels - {DROC}
  [deutsches {ROman} corpus]}.

\bibitem[{Leblond(2022)}]{Leblond_2022}
Aude Leblond. 2022.
\newblock \href {https://zenodo.org/records/7446728} {Corpus chapitres}.
\newblock ANR Chapitres.

\bibitem[{Lee et~al.(2017)Lee, He, Lewis, and Zettlemoyer}]{Lee_2017}
Kenton Lee, Luheng He, Mike Lewis, and Luke Zettlemoyer. 2017.
\newblock \href {https://doi.org/10.18653/v1/D17-1018} {End-to-end neural
  coreference resolution}.
\newblock In \emph{Proceedings of the 2017 Conference on Empirical Methods in
  Natural Language Processing}, pages 188--197, Copenhagen, Denmark.
  Association for Computational Linguistics.

\bibitem[{Lee et~al.(2018)Lee, He, and Zettlemoyer}]{Lee_2018}
Kenton Lee, Luheng He, and Luke Zettlemoyer. 2018.
\newblock \href {https://doi.org/10.18653/v1/N18-2108} {Higher-order
  coreference resolution with coarse-to-fine inference}.
\newblock In \emph{Proceedings of the 2018 Conference of the North {A}merican
  Chapter of the Association for Computational Linguistics: Human Language
  Technologies, Volume 2 (Short Papers)}, pages 687--692, New Orleans,
  Louisiana. Association for Computational Linguistics.

\bibitem[{Liu et~al.(2021)Liu, Shi, and Chen}]{liu_2021}
Zhengyuan Liu, Ke~Shi, and Nancy Chen. 2021.
\newblock \href {https://doi.org/10.18653/v1/2021.sigdial-1.53}
  {Coreference-aware dialogue summarization}.
\newblock In \emph{Proceedings of the 22nd Annual Meeting of the Special
  Interest Group on Discourse and Dialogue}, pages 509--519, Singapore and
  Online. Association for Computational Linguistics.

\bibitem[{Lu and Poesio(2021)}]{Lu_2021}
Pengcheng Lu and Massimo Poesio. 2021.
\newblock \href {https://doi.org/10.18653/v1/2021.crac-1.2} {Coreference
  resolution for the biomedical domain: A survey}.
\newblock In \emph{Proceedings of the Fourth Workshop on Computational Models
  of Reference, Anaphora and Coreference}, pages 12--23, Punta Cana, Dominican
  Republic. Association for Computational Linguistics.

\bibitem[{Luo(2005)}]{Luo_2005}
Xiaoqiang Luo. 2005.
\newblock \href {https://aclanthology.org/H05-1004/} {On coreference resolution
  performance metrics}.
\newblock In \emph{Proceedings of Human Language Technology Conference and
  Conference on Empirical Methods in Natural Language Processing}, pages
  25--32, Vancouver, British Columbia, Canada. Association for Computational
  Linguistics.

\bibitem[{Martin et~al.(2020)Martin, Muller, Ortiz~Su{\'a}rez, Dupont, Romary,
  de~la Clergerie, Seddah, and Sagot}]{Martin_2020}
Louis Martin, Benjamin Muller, Pedro~Javier Ortiz~Su{\'a}rez, Yoann Dupont,
  Laurent Romary, {\'E}ric de~la Clergerie, Djam{\'e} Seddah, and Beno{\^i}t
  Sagot. 2020.
\newblock \href {https://doi.org/10.18653/v1/2020.acl-main.645} {{C}amem{BERT}:
  a tasty {F}rench language model}.
\newblock In \emph{Proceedings of the 58th Annual Meeting of the Association
  for Computational Linguistics}, pages 7203--7219, Online. Association for
  Computational Linguistics.

\bibitem[{Martinelli et~al.(2024)Martinelli, Barba, and
  Navigli}]{Martinelli_2024}
Giuliano Martinelli, Edoardo Barba, and Roberto Navigli. 2024.
\newblock \href {https://doi.org/10.18653/v1/2024.acl-long.722} {Maverick:
  Efficient and accurate coreference resolution defying recent trends}.
\newblock In \emph{Proceedings of the 62nd Annual Meeting of the Association
  for Computational Linguistics (Volume 1: Long Papers)}, pages 13380--13394,
  Bangkok, Thailand. Association for Computational Linguistics.

\bibitem[{Martinelli et~al.(2025)Martinelli, Bonomo, Huguet~Cabot, and
  Navigli}]{martinelli_2025}
Giuliano Martinelli, Tommaso Bonomo, Pere-Llu{\'i}s Huguet~Cabot, and Roberto
  Navigli. 2025.
\newblock \href {https://doi.org/10.18653/v1/2025.acl-long.1197} {{BOOKCOREF}:
  Coreference resolution at book scale}.
\newblock In \emph{Proceedings of the 63rd Annual Meeting of the Association
  for Computational Linguistics (Volume 1: Long Papers)}, pages 24526--24544,
  Vienna, Austria. Association for Computational Linguistics.

\bibitem[{Moosavi and Strube(2016)}]{Moosavi_2016}
Nafise~Sadat Moosavi and Michael Strube. 2016.
\newblock \href {https://doi.org/10.18653/v1/P16-1060} {Which coreference
  evaluation metric do you trust? a proposal for a link-based entity aware
  metric}.
\newblock In \emph{Proceedings of the 54th Annual Meeting of the Association
  for Computational Linguistics (Volume 1: Long Papers)}, pages 632--642,
  Berlin, Germany. Association for Computational Linguistics.

\bibitem[{Moretti(2013)}]{Moretti_2013}
Franco Moretti. 2013.
\newblock \emph{Distant Reading}.
\newblock Verso, London.

\bibitem[{Muzny et~al.(2017)Muzny, Fang, Chang, and Jurafsky}]{Muzny_2017}
Grace Muzny, Michael Fang, Angel Chang, and Dan Jurafsky. 2017.
\newblock \href {https://aclanthology.org/E17-1044/} {A two-stage sieve
  approach for quote attribution}.
\newblock In \emph{Proceedings of the 15th Conference of the {E}uropean Chapter
  of the Association for Computational Linguistics: Volume 1, Long Papers},
  pages 460--470, Valencia, Spain. Association for Computational Linguistics.

\bibitem[{Mélanie et~al.(2024)Mélanie, Barré, Seminck, Plancq, Naguib,
  Pastor, and Poibeau}]{Melanie_2024}
Frédérique Mélanie, Jean Barré, Olga Seminck, Clément Plancq, Marco
  Naguib, Martial Pastor, and Thierry Poibeau. 2024.
\newblock \href {https://doi.org/10.48694/jcls.3924} {Booknlp-fr, the french
  versant of booknlp. a tailored pipeline for 19th and 20th century french
  literature}.
\newblock \emph{Journal of Computational Literary Studies}, 3(1):1--34.

\bibitem[{Naguib et~al.(2022)Naguib, Delaborde, Andrault, Bekolo, and
  Seminck}]{Naguib_2022}
Marco Naguib, Marine Delaborde, Blandine Andrault, Ana{\"i}s Bekolo, and Olga
  Seminck. 2022.
\newblock \href {https://aclanthology.org/2022.jeptalnrecital-humanum.8/}
  {Romanciers et romanci{\`e}res du {XIX}{\`e}me si{\`e}cle : une {\'e}tude
  automatique du genre sur le corpus {GIRLS} (male and female novelists : an
  automatic study of gender of authors and their characters )}.
\newblock In \emph{Actes de la 29e Conf{\'e}rence sur le Traitement Automatique
  des Langues Naturelles. Atelier TAL et Humanit{\'e}s Num{\'e}riques
  (TAL-HN)}, pages 66--77, Avignon, France. ATALA.

\bibitem[{Oberle(2018)}]{Oberle_2018}
Bruno Oberle. 2018.
\newblock Sacr: A drag-and-drop based tool for coreference annotation.
\newblock In \emph{Proceedings of the 11th Edition of the Language Resources
  and Evaluation Conference (LREC 2018)}, Miyazaki, Japan.

\bibitem[{O{'}Keefe et~al.(2013)O{'}Keefe, Webster, Curran, and
  Koprinska}]{OKeefe_2013}
Tim O{'}Keefe, Kellie Webster, James~R. Curran, and Irena Koprinska. 2013.
\newblock \href {https://aclanthology.org/U13-1007/} {Examining the impact of
  coreference resolution on quote attribution}.
\newblock In \emph{Proceedings of the Australasian Language Technology
  Association Workshop 2013 ({ALTA} 2013)}, pages 43--52, Brisbane, Australia.

\bibitem[{Otmazgin et~al.(2022)Otmazgin, Cattan, and Goldberg}]{Otmazgin_2022}
Shon Otmazgin, Arie Cattan, and Yoav Goldberg. 2022.
\newblock \href {https://doi.org/10.18653/v1/2022.aacl-demo.6} {{F}-coref:
  Fast, accurate and easy to use coreference resolution}.
\newblock In \emph{Proceedings of the 2nd Conference of the Asia-Pacific
  Chapter of the Association for Computational Linguistics and the 12th
  International Joint Conference on Natural Language Processing: System
  Demonstrations}, pages 48--56, Taipei, Taiwan. Association for Computational
  Linguistics.

\bibitem[{Poesio et~al.(2023)Poesio, Yu, Paun, Aloraini, Lu, Haber, and
  Cokal}]{Poesio_2023}
Massimo Poesio, Juntao Yu, Silviu Paun, Abdulrahman Aloraini, Pengcheng Lu,
  Janosch Haber, and Derya Cokal. 2023.
\newblock \href {https://doi.org/10.1146/annurev-linguistics-031120-111653}
  {Computational models of anaphora}.
\newblock \emph{Annual Review of Linguistics}, 9(Volume 9, 2023):561--587.

\bibitem[{Poot and van Cranenburgh(2020)}]{Poot_2020}
Corb{\`e}n Poot and Andreas van Cranenburgh. 2020.
\newblock \href {https://aclanthology.org/2020.crac-1.9/} {A benchmark of
  rule-based and neural coreference resolution in {D}utch novels and news}.
\newblock In \emph{Proceedings of the Third Workshop on Computational Models of
  Reference, Anaphora and Coreference}, pages 79--90, Barcelona, Spain
  (online). Association for Computational Linguistics.

\bibitem[{Porada et~al.(2024)Porada, Zou, and Cheung}]{porada_2024}
Ian Porada, Xiyuan Zou, and Jackie Chi~Kit Cheung. 2024.
\newblock \href {https://aclanthology.org/2024.lrec-main.23/} {A controlled
  reevaluation of coreference resolution models}.
\newblock In \emph{Proceedings of the 2024 Joint International Conference on
  Computational Linguistics, Language Resources and Evaluation (LREC-COLING
  2024)}, pages 256--263, Torino, Italia. ELRA and ICCL.

\bibitem[{Pradhan et~al.(2012)Pradhan, Moschitti, Xue, Uryupina, and
  Zhang}]{Pradhan_2012}
Sameer Pradhan, Alessandro Moschitti, Nianwen Xue, Olga Uryupina, and Yuchen
  Zhang. 2012.
\newblock \href {https://aclanthology.org/W12-4501/} {{C}o{NLL}-2012 shared
  task: Modeling multilingual unrestricted coreference in {O}nto{N}otes}.
\newblock In \emph{Joint Conference on {EMNLP} and {C}o{NLL} - Shared Task},
  pages 1--40, Jeju Island, Korea. Association for Computational Linguistics.

\bibitem[{Pradhan et~al.(2011)Pradhan, Ramshaw, Marcus, Palmer, Weischedel, and
  Xue}]{Pradhan_2011}
Sameer Pradhan, Lance Ramshaw, Mitchell Marcus, Martha Palmer, Ralph
  Weischedel, and Nianwen Xue. 2011.
\newblock \href {https://aclanthology.org/W11-1901/} {{C}o{NLL}-2011 shared
  task: Modeling unrestricted coreference in {O}nto{N}otes}.
\newblock In \emph{Proceedings of the Fifteenth Conference on Computational
  Natural Language Learning: Shared Task}, pages 1--27, Portland, Oregon, USA.
  Association for Computational Linguistics.

\bibitem[{Recasens and Hovy(2011)}]{Recasens_2011}
M.~Recasens and E.~Hovy. 2011.
\newblock \href {https://doi.org/10.1017/S135132491000029X} {Blanc:
  Implementing the rand index for coreference evaluation}.
\newblock \emph{Nat. Lang. Eng.}, 17(4):485–510.

\bibitem[{Roesiger et~al.(2018)Roesiger, Schulz, and Reiter}]{Roesiger_2018}
Ina Roesiger, Sarah Schulz, and Nils Reiter. 2018.
\newblock \href {https://aclanthology.org/W18-4515/} {Towards coreference for
  literary text: Analyzing domain-specific phenomena}.
\newblock In \emph{Proceedings of the Second Joint {SIGHUM} Workshop on
  Computational Linguistics for Cultural Heritage, Social Sciences, Humanities
  and Literature}, pages 129--138, Santa Fe, New Mexico. Association for
  Computational Linguistics.

\bibitem[{Shridhar et~al.(2023)Shridhar, Monath, Thirukovalluru, Stolfo,
  Zaheer, McCallum, and Sachan}]{Shridhar_2023}
Kumar Shridhar, Nicholas Monath, Raghuveer Thirukovalluru, Alessandro Stolfo,
  Manzil Zaheer, Andrew McCallum, and Mrinmaya Sachan. 2023.
\newblock \href {https://doi.org/10.18653/v1/2023.findings-eacl.105}
  {Longtonotes: {O}nto{N}otes with longer coreference chains}.
\newblock In \emph{Findings of the Association for Computational Linguistics:
  EACL 2023}, pages 1428--1442, Dubrovnik, Croatia. Association for
  Computational Linguistics.

\bibitem[{Thirukovalluru et~al.(2021)Thirukovalluru, Monath, Shridhar, Zaheer,
  Sachan, and McCallum}]{Thirukovalluru_2021}
Raghuveer Thirukovalluru, Nicholas Monath, Kumar Shridhar, Manzil Zaheer,
  Mrinmaya Sachan, and Andrew McCallum. 2021.
\newblock \href {https://doi.org/10.18653/v1/2021.findings-acl.343} {Scaling
  within document coreference to long texts}.
\newblock In \emph{Findings of the Association for Computational Linguistics:
  ACL-IJCNLP 2021}, pages 3921--3931, Online. Association for Computational
  Linguistics.

\bibitem[{Toshniwal et~al.(2020)Toshniwal, Wiseman, Ettinger, Livescu, and
  Gimpel}]{Toshniwal_2020}
Shubham Toshniwal, Sam Wiseman, Allyson Ettinger, Karen Livescu, and Kevin
  Gimpel. 2020.
\newblock \href {https://doi.org/10.18653/v1/2020.emnlp-main.685} {Learning to
  {I}gnore: {L}ong {D}ocument {C}oreference with {B}ounded {M}emory {N}eural
  {N}etworks}.
\newblock In \emph{Proceedings of the 2020 Conference on Empirical Methods in
  Natural Language Processing (EMNLP)}, pages 8519--8526, Online. Association
  for Computational Linguistics.

\bibitem[{Underwood et~al.(2018)Underwood, Bamman, and Lee}]{Underwood_2018}
Ted Underwood, David Bamman, and Sabrina Lee. 2018.
\newblock \href {https://doi.org/10.22148/16.019} {The transformation of gender
  in english-language fiction}.
\newblock \emph{Cultural Analytics}.

\bibitem[{van Cranenburgh(2019)}]{Cranenburgh_2019}
Andreas van Cranenburgh. 2019.
\newblock \href {https://clinjournal.org/clinj/article/view/91} {A dutch
  coreference resolution system with an evaluation on literary fiction}.
\newblock \emph{Computational Linguistics in the Netherlands Journal},
  9:27–54.

\bibitem[{{van Zundert} et~al.(2023){van Zundert}, {van Cranenburgh}, and
  Smeets}]{Zundert_2023}
Joris {van Zundert}, Andreas {van Cranenburgh}, and Roel Smeets. 2023.
\newblock Putting dutchcoref to the test: Character detection and gender
  dynamics in contemporary dutch novels.
\newblock In \emph{Proceedings of the Computational Humanities Research
  conference 2023}, pages 757--771. CEUR Workshop Proceedings (CEUR-WS.org).
\newblock Computational Humanities Research Conference ; Conference date:
  06-12-2023 Through 08-12-2023.

\bibitem[{Varoquaux et~al.(2025)Varoquaux, Sasha~Luccioni, and
  Whittaker}]{Varoquaux_2025}
Gaël Varoquaux, Alexandra Sasha~Luccioni, and Meredith Whittaker. 2025.
\newblock \href {https://arxiv.org/abs/2409.14160} {Hype, sustainability, and
  the price of the bigger-is-better paradigm in ai}.
\newblock \emph{Preprint}, arXiv:2409.14160.

\bibitem[{Vianne et~al.(2023)Vianne, Dupont, and Barr{\'e}}]{Vianne_2023}
Laurine Vianne, Yoann Dupont, and Jean Barr{\'e}. 2023.
\newblock \href {https://hal.science/hal-04447642} {{Gender Bias in French
  Literature}}.
\newblock In \emph{{Conference on Computational Humanities Research CHR2023}},
  Paris, France. {Ariane and Epita and Humanistica}.

\bibitem[{Vilain et~al.(1995)Vilain, Burger, Aberdeen, Connolly, and
  Hirschman}]{Vilain_1995}
Marc~B. Vilain, John~D. Burger, John~S. Aberdeen, Dennis Connolly, and Lynette
  Hirschman. 1995.
\newblock \href {https://doi.org/10.3115/1072399.1072405} {A model-theoretic
  coreference scoring scheme}.
\newblock In \emph{Proceedings of the 6th Conference on Message Understanding,
  {MUC} 1995, Columbia, Maryland, USA, November 6-8, 1995}, pages 45--52.
  {ACL}.

\bibitem[{Vishnubhotla et~al.(2022)Vishnubhotla, Hammond, and
  Hirst}]{Vishnubhotla_2022}
Krishnapriya Vishnubhotla, Adam Hammond, and Graeme Hirst. 2022.
\newblock \href {https://aclanthology.org/2022.lrec-1.628/} {The project
  dialogism novel corpus: A dataset for quotation attribution in literary
  texts}.
\newblock In \emph{Proceedings of the Thirteenth Language Resources and
  Evaluation Conference}, pages 5838--5848, Marseille, France. European
  Language Resources Association.

\bibitem[{Vishnubhotla et~al.(2023)Vishnubhotla, Rudzicz, Hirst, and
  Hammond}]{Vishnubhotla_2023}
Krishnapriya Vishnubhotla, Frank Rudzicz, Graeme Hirst, and Adam Hammond. 2023.
\newblock \href {https://doi.org/10.18653/v1/2023.acl-short.64} {Improving
  automatic quotation attribution in literary novels}.
\newblock In \emph{Proceedings of the 61st Annual Meeting of the Association
  for Computational Linguistics (Volume 2: Short Papers)}, pages 737--746,
  Toronto, Canada. Association for Computational Linguistics.

\bibitem[{Vu et~al.(2024)Vu, Kamigaito, and Watanabe}]{vu_2024}
Huy~Hien Vu, Hidetaka Kamigaito, and Taro Watanabe. 2024.
\newblock \href {https://doi.org/10.1162/tacl_a_00677} {Context-aware machine
  translation with source coreference explanation}.
\newblock \emph{Transactions of the Association for Computational Linguistics},
  12:856--874.

\bibitem[{Weischedel et~al.(2013)Weischedel, Palmer, Marcus, Hovy, Pradhan,
  Ramshaw, Xue, Taylor, Kaufman, Franchini, El-Bachouti, Belvin, and
  Houston}]{Weischedel_2013}
Ralph Weischedel, Martha Palmer, Mitchell Marcus, Eduard Hovy, Sameer Pradhan,
  Lance Ramshaw, Nianwen Xue, Ann Taylor, Jeff Kaufman, Michelle Franchini,
  Mohammed El-Bachouti, Robert Belvin, and Ann Houston. 2013.
\newblock {OntoNotes} release 5.0.

\bibitem[{Wiseman et~al.(2015)Wiseman, Rush, Shieber, and
  Weston}]{Wiseman_2015}
Sam Wiseman, Alexander~M. Rush, Stuart Shieber, and Jason Weston. 2015.
\newblock \href {https://doi.org/10.3115/v1/P15-1137} {Learning anaphoricity
  and antecedent ranking features for coreference resolution}.
\newblock In \emph{Proceedings of the 53rd Annual Meeting of the Association
  for Computational Linguistics and the 7th International Joint Conference on
  Natural Language Processing (Volume 1: Long Papers)}, pages 1416--1426,
  Beijing, China. Association for Computational Linguistics.

\bibitem[{Wu et~al.(2020)Wu, Wang, Yuan, Wu, and Li}]{Wu_2020}
Wei Wu, Fei Wang, Arianna Yuan, Fei Wu, and Jiwei Li. 2020.
\newblock \href {https://doi.org/10.18653/v1/2020.acl-main.622} {{C}oref{QA}:
  Coreference resolution as query-based span prediction}.
\newblock In \emph{Proceedings of the 58th Annual Meeting of the Association
  for Computational Linguistics}, pages 6953--6963, Online. Association for
  Computational Linguistics.

\bibitem[{Xu and Choi(2020)}]{Xu_2020}
Liyan Xu and Jinho~D. Choi. 2020.
\newblock \href {https://doi.org/10.18653/v1/2020.emnlp-main.686} {Revealing
  the myth of higher-order inference in coreference resolution}.
\newblock In \emph{Proceedings of the 2020 Conference on Empirical Methods in
  Natural Language Processing (EMNLP)}, pages 8527--8533, Online. Association
  for Computational Linguistics.

\bibitem[{Yao et~al.(2019)Yao, Ye, Li, Han, Lin, Liu, Liu, Huang, Zhou, and
  Sun}]{yao_2019}
Yuan Yao, Deming Ye, Peng Li, Xu~Han, Yankai Lin, Zhenghao Liu, Zhiyuan Liu,
  Lixin Huang, Jie Zhou, and Maosong Sun. 2019.
\newblock \href {https://doi.org/10.18653/v1/P19-1074} {{D}oc{RED}: A
  large-scale document-level relation extraction dataset}.
\newblock In \emph{Proceedings of the 57th Annual Meeting of the Association
  for Computational Linguistics}, pages 764--777, Florence, Italy. Association
  for Computational Linguistics.

\end{thebibliography}

\newpage
\appendix

\section{Mention Detection Model}
\label{sec:mention_detection_model_details}

The mention detection module consists of two stacked BiLSTM-CRF models, each trained on a different nesting level of mentions. During inference, predicted spans from both models are combined. If two mention spans overlap, the span with the lower prediction confidence is discarded.

\textbf{BERT embeddings}: The raw text is split into overlapping segments of length $L$ (the maximum embedding model context window) with an overlap of $L/2$ to maximize the context available for each token. Each segment is passed through the CamemBERT\textsubscript{LARGE} model, and we retrieve the last hidden layer as the token representations (1024 dimensions). The final token embedding is computed as the average from overlapping segments. We do not fine-tune CamemBERT for this task.

\textbf{BIOES tag prediction}: For each sentence, token representations are passed through the BiLSTM-CRF model, which outputs a sequence of BIOES tags:  
B-PER (Beginning of mention), I-PER (Inside), E-PER (End), S-PER (Single-token mention), and O (Outside).

\subsection{Model Architecture}
\begin{itemize}[leftmargin=10pt, itemsep=1pt, topsep=1pt]  
    \item \textbf{Locked Dropout} (0.5) applied to embeddings for regularization.
    \item \textbf{Projection Layer}: Highway network mapping 1024 $\to$ 2048 dimensions.
    \item \textbf{BiLSTM Layer}: Single bidirectional LSTM (256 hidden units per direction).
    \item \textbf{Linear Layer}: Maps 512-dimensional BiLSTM outputs to BIOES label scores.
    \item \textbf{CRF Layer}: Enforces structured consistency in predictions.
\end{itemize}

\subsection{Model Training} 
\begin{itemize}[leftmargin=10pt, itemsep=1pt, topsep=1pt]  
    \item \textbf{Data Splitting}: Leave-One-Out Cross-Validation (LOOCV) with an 85\%/15\% train-validation split.  
    \item \textbf{Batch Size}: 16 sentences per batch.  
    \item \textbf{Optimization}: Adam optimizer (lr = $1.4 \times 10^{-4}$, weight decay = $10^{-5}$).  
    \item \textbf{Learning Rate Scheduling}: ReduceLROnPlateau (factor = 0.5, patience = 2).  
    \item \textbf{Average Training Epochs}: 20.  
    \item \textbf{Hardware}: Trained on a single 6GB Nvidia RTX 1000 Ada Generation GPU.  
\end{itemize}

\section{Nearest Antecedent Distribution}
\label{sec:antecedent_distribution}

\begin{figure}[htbp]
\vspace{-0.5em}  
  \centering
  \includegraphics[width=\columnwidth]{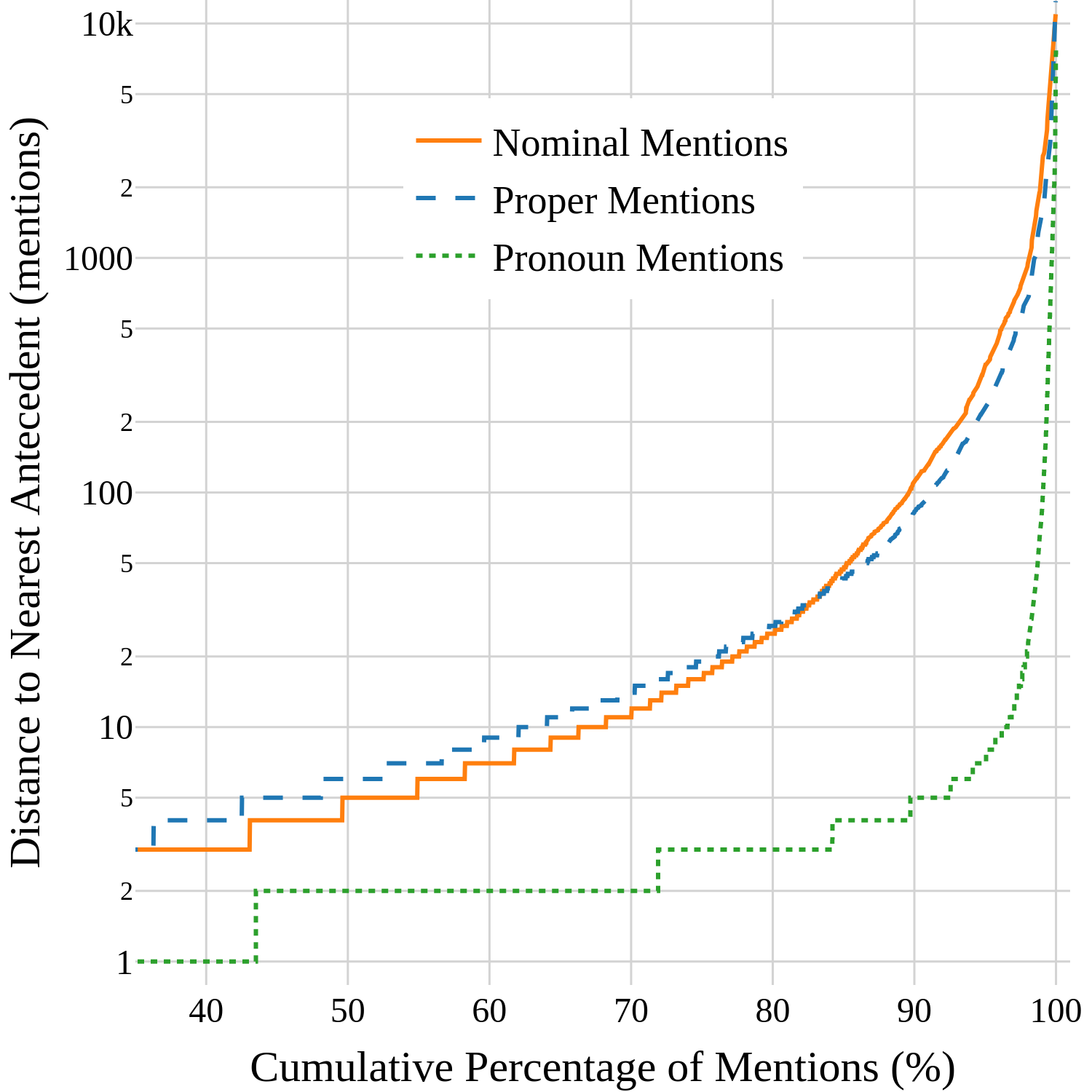}  
  \vspace{-1em}  
  \caption{Distance to nearest antecedent for mentions of different type.}
  \vspace{-1em}  
  \label{fig:nearest_antecedent_distribution}
\end{figure}

\section{Coreference Resolution Model}
\label{sec:mention_pairs_model}

\subsection{Model Architecture}
\begin{itemize}[leftmargin=10pt, itemsep=1pt, topsep=1pt]  
    \item \textbf{Model Input}: 2,165-dimensional vector, composed of concatenated:  
    \begin{itemize}[leftmargin=10pt, itemsep=1pt, topsep=1pt]  
        \item \textbf{CamemBERT embeddings}: Maximum context embeddings for both mentions (2 × 1,024 = 2,048 dimensions).  
        \item \textbf{Mention Features} (106 dimensions):  
        \begin{itemize}[leftmargin=10pt, itemsep=1pt, topsep=1pt]
            \item Mention length.  
            \item Position of the mention's start token in the sentence.  
            \item Grammatical category (pronoun, common noun, proper noun).  
            \item Dependency relation of the mention's head (one-hot encoded).  
            \item Gender (one-hot encoded).  
            \item Number (one-hot encoded).  
            \item Grammatical person (one-hot encoded).  
        \end{itemize}  
        \item \textbf{Mention Pair Features} (11 dimensions):  
        \begin{itemize}[leftmargin=10pt, itemsep=1pt, topsep=1pt]
            \item Distance between mention IDs.  
            \item Distance between start and end tokens of mentions.  
            \item Sentence and paragraph distance.  
            \item Difference in nesting levels.  
            \item Ratio of shared tokens between mentions.  
            \item Exact text match (binary).  
            \item Exact match of mention heads (binary).  
            \item Match of syntactic heads (binary).  
            \item Match of entity types (binary).  
        \end{itemize}  
    \end{itemize}  

    \item \textbf{Hidden Layers}:  
    \begin{itemize}[leftmargin=10pt, itemsep=1pt, topsep=1pt]
        \item Three fully connected layers.  
        \item 1,900 hidden units per layer with ReLU activation.  
        \item Dropout rate of 0.6 for regularization.  
    \end{itemize}  

    \item \textbf{Final Layer}:  
    \begin{itemize}[leftmargin=10pt, itemsep=1pt, topsep=1pt]
        \item Linear layer mapping from 1,900 dimensions to a single scalar score.  
        \item Output: Continuous value between 0 (not coreferent) and 1 (coreferent).  
    \end{itemize}  
\end{itemize}  

\subsection{Model Training}  
\begin{itemize}[leftmargin=10pt, itemsep=1pt, topsep=1pt]  
    \item \textbf{Data Splitting}: Leave-One-Out Cross-Validation (LOOCV) with an 85\%/15\% train-validation split.  
    \item \textbf{Batch Size}: 16,000 mention-pairs per batch.  
    \item \textbf{Optimization}: Adam optimizer (lr = $1.4 \times 10^{-4}$, weight decay = $10^{-5}$).
    \item \textbf{Antecedent Candidates}:  
    \begin{itemize}[leftmargin=10pt, itemsep=1pt, topsep=1pt]
        \item 30 for pronouns.  
        \item 300 for common and proper nouns.  
    \end{itemize}
    \item \textbf{Hardware}: Trained on a single 6GB Nvidia RTX 1000 Ada Generation GPU.  
\end{itemize}

\section{Mention-Pairs Scorer Error Distribution}
\label{sec:mention_pairs_errors}

\begin{figure}[htbp]
\vspace{-0.5em}  
  \centering
  \includegraphics[width=\columnwidth]{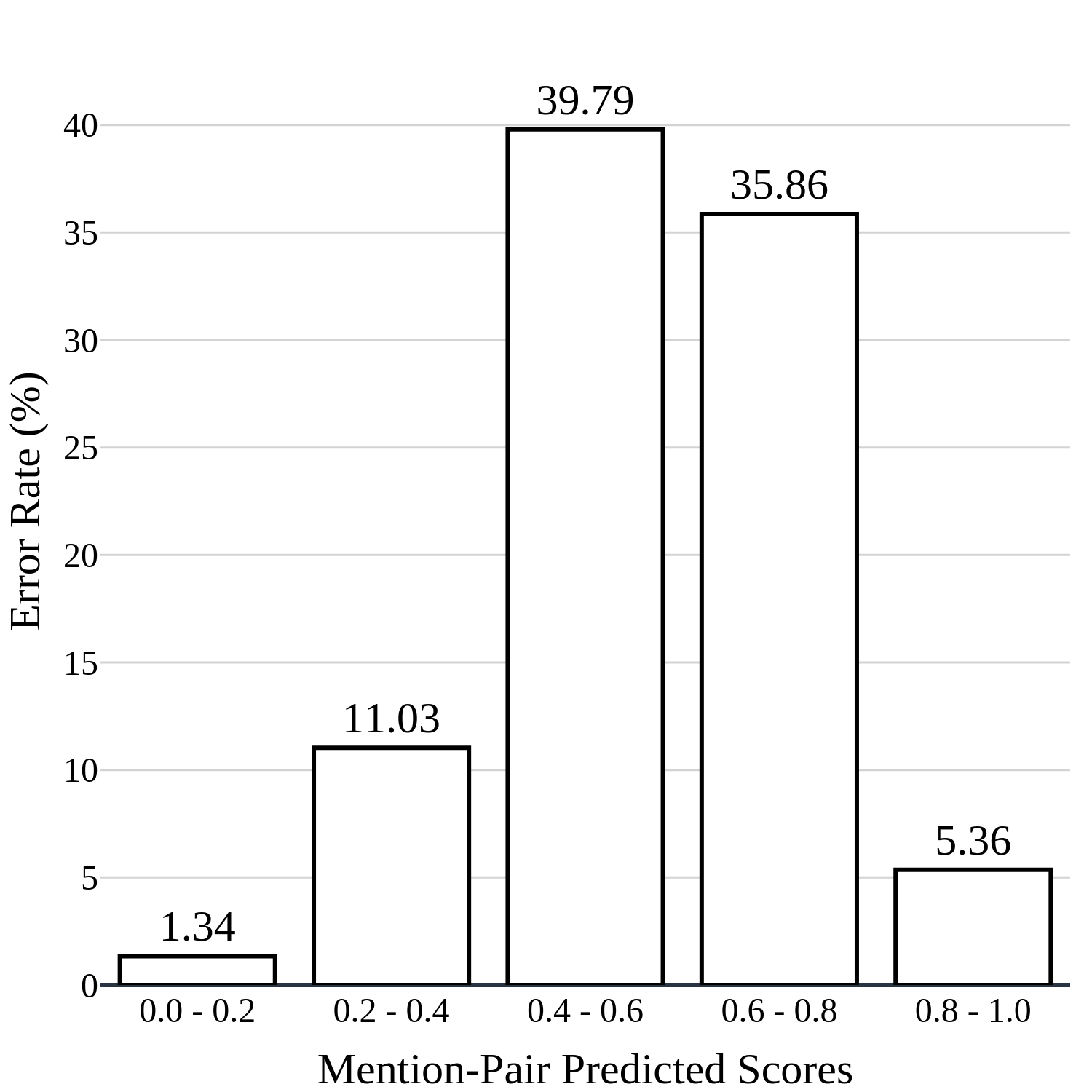}  
  \vspace{-1em}  
  \caption{Error Rate by Mention-pair Predicted Score Range.}
  \vspace{-1em}  
  \label{fig:mention_pairs_scorer_errors}
\end{figure}

\begin{table*}[htbp]
  \centering
  \noindent\section{Detailed performance gain from clustering strategy}
  \label{sec:absolute_gain_clustering_strategy}
  \includegraphics[width=\textwidth]{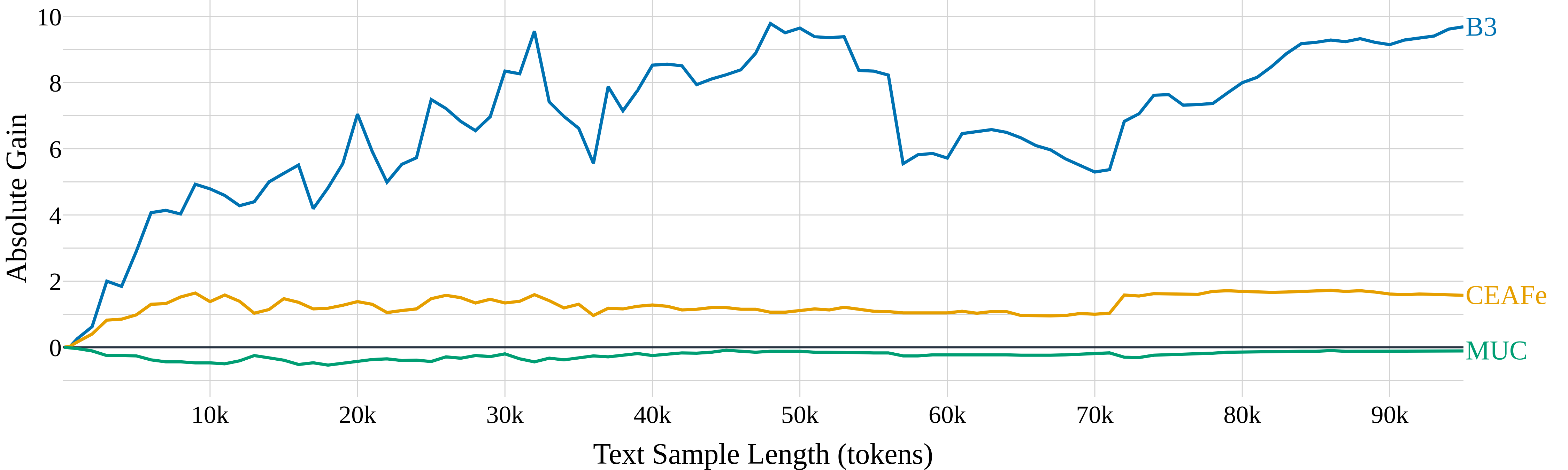}  
  \vspace{-1.5em}  
  \caption{Absolute CR performance gain from the global proper mentions clustering strategy over vanilla left-to-right, as a function of document length. Predicted mentions.}
  \vspace{-1.em}  
  \label{fig:performance_gain_function_of_document_length_figure}
  \centering
  \label{tab:absolute_gain_clustering_strategy}
\end{table*}

\begin{table*}[htbp]
  \centering
  \noindent\section{Annotated Dataset Details}
  \label{sec:Annotated_Dataset_Details}
  \centering
  \setlength{\tabcolsep}{27pt}
  \begin{tabular}{lccc}
    \hline
    \textbf{Year} & \textbf{Author} & \textbf{Text} & \textbf{Tokens} \\
    \hline
    1731 & Antoine-François Prévost & \textit{Manon Lescaut} & 71,219 \\
    1832 & George Sand & \textit{Indiana} & 115,415 \\
    1923 & Delly & \textit{Dans les ruines} & 98,542 \\
    \hline
  \end{tabular}
  \caption{Annotated Dataset Details}
  \label{tab:corpus}
\end{table*}

\begin{table*}[htbp]
  \centering

  \noindent\section{Comparison of CR performance with other datasets and languages}
  \label{sec:coref_benchmarks_comparison}
  \centering
  \setlength{\tabcolsep}{0pt}
  \renewcommand{\arraystretch}{1.1}
  \small
  {\fontsize{9.5}{12}\selectfont
    \begin{tabular}{@{}p{3.7cm} >{\centering\arraybackslash}p{3.6cm} >{\centering\arraybackslash}p{1.6cm} >{\centering\arraybackslash}p{2.1cm} >{\centering\arraybackslash}p{1.2cm} >{\centering\arraybackslash}p{1.2cm} >{\centering\arraybackslash}p{1.3cm}>{\centering\arraybackslash}p{1.3cm}@{}}
        \hline
        \textbf{Corpus} & \textbf{Model} & \textbf{Mentions} & \textbf{Tokens / Doc} & \textbf{MUC} & \textbf{B\textsuperscript{3}} & \textbf{CEAFe} & \textbf{CoNLL} \\
        \hline
        LitBank (\textit{English}) & \citealt{Bamman_2020} & Gold & 2,105 & 88.5 & 72.6 & 76.7 & 79.3 \\
        LitBank-fr (LOOCV) & Ours & Gold & 2,105 & 91.93 & 74.6 & 75.35 & 80.63 \\
        \hline
        LitBank (\textit{English}) & \citealt{Bamman_2020} & Predicted & 2,105 & 84.3 & 62.73 & 57.3 & 68.1 \\
        LitBank (\textit{English}) & \citealt{Thirukovalluru_2021} & Predicted & 2,105 & 89.50 & 78.21 & 67.59 & 78.44 \\
        LitBank-fr (LOOCV) & Ours & Predicted & 2,105 & 84.58 & 74.77 & 63.30 & 73.21 \\
        \hline
        KoCoNovel (\textit{Korean}) & \citealt{Kim_2024} & Predicted & 3,578 & 71.06 & 57.33 & 44.19 & 57.53 \\
        Long-LitBank-fr (LOOCV) & Ours & Predicted & 3,578 & 88.31 & 68.79 & 47.17 & 68.09 \\
        \hline
        G. Orwell, \textit{Animal Farm} & \citealt{Guo_2023} & Predicted & 37,000 & - & - & - & 36.3 \\
        Long-LitBank-fr (LOOCV) & Ours & Predicted & 37,000 & 92.79 & 52.35 & 32.89 & 59.34 \\
        \hline
        BookCoref\textsubscript{gold} & Longdoc & Predicted & 76,419 & 93.5 & 62.4 & 45.3 & 67.0 \\
        BookCoref\textsubscript{gold} & Maverick\textsubscript{xl} & Predicted & 76,419 & 94.3 & 55.3 & 33.4 & 61.0 \\
        Long-LitBank-fr (LOOCV) & Ours & Predicted & 76,000 & 94.99 & 47.51 & 37.49 & 60.00 \\
        \hline
      \end{tabular}}
      \caption{Comparison of CR performance with other work on literary coreference with predicted and gold mentions.}
\end{table*}

\section*{}
\section*{}

\end{document}